\documentclass{article}

\usepackage[preprint]{neurips_2026}

\usepackage[utf8]{inputenc} % allow utf-8 input
\usepackage[T1]{fontenc}    % use 8-bit T1 fonts
\usepackage{hyperref}       % hyperlinks
\usepackage{url}            % simple URL typesetting
\usepackage{booktabs}       % professional-quality tables
\usepackage{amsfonts}       % blackboard math symbols
\usepackage{nicefrac}       % compact symbols for 1/2, etc.
\usepackage{microtype}      % microtypography
\usepackage{xcolor}         % colors
\usepackage{ulem}
\usepackage{graphicx}
\usepackage{amssymb}% http://ctan.org/pkg/amssymb
\usepackage{pifont}% http://ctan.org/pkg/pifont
\usepackage{wrapfig}
\usepackage{amsmath}
\usepackage{multirow}
\usepackage{enumitem}

\newcommand{\cmark}{\ding{51}}%
\newcommand{\xmark}{\ding{55}}%

\def\modelname~{RoboSynChallenge}

\title{RoboSynChallenge: Mastering Real-World Dexterity via Generalizing Synthesized Manipulation Skills}

\author{%
  Runyi Zhao$^{1,2}$,
  Ruixin Wu$^{1,2}$,
  Chengkun Li$^{1,2}$,
  Hongrui Zhang$^{1,2}$, \\
  Ang Li$^{2}$,
  Ruixing Jin$^{2}$,
  Yueci Deng$^{2,11}$,
  Yingying Guo$^{2}$, \\
  Lihe Ding$^{3}$,
  Shaocong Dong$^{4}$,
  Tianfan Xue$^{3}$,
  Yanjun Gao$^{5}$,
  Yudong Luo$^{6}$,\\
  Pascal Poupart$^{7,8}$,
  Simo Wu$^{9}$,
  Kui Jia$^{2,11}$,
  Wei-shi Zheng$^{1,10}$,
  Guiliang Liu$^{1,2}$ \\[0.15cm]
  \small $^1$Shenzhen Loop Area Institute (SLAI) \quad $^2$ The Chinese University of Hong Kong, Shenzhen \\
  \small $^3$ The Chinese University of Hong Kong \quad
  \small $^4$ The Hong Kong University of Science and Technology \\
  \small $^5$ LARK Lab, University of Colorado Anschutz \quad
  \small $^6$ Mila - Quebec AI Institute, Canada \quad 
  \small $^7$ Vector Institute \\
  \small $^8$ University of Waterloo \quad 
  \small $^9$ Fudan University \quad 
  \small $^{10}$ Sun Yat-sen University \quad 
  \small $^{11}$ DexForce\\[0.25cm] 
{\tt robosynchallenge@gmail.com}}

\begin{document}

\maketitle
\vspace{-0.25in}
% \vspace{-0.05in}

\begin{center}
\small
\ding{108}~\textbf{Website}: \url{https://robosyn-bench.net/}\\
\ding{117}~\textbf{GitHub}: \url{https://github.com/EDEM-AI/RoboSynChallenge/}\\
\ding{45}~\textbf{Tutorial}: \url{https://edem-ai.github.io/RoboSynChallenge/html/}
\end{center}
\vspace{0.1in}

\begin{abstract}
Achieving generalizable robotic manipulation remains a central challenge in embodied intelligence. Despite rapid advances in model architectures and learning algorithms, progress is often limited by the scarcity and narrow diversity of real-world data. The \textbf{RoboSynChallenge} competition introduces a unified benchmark to evaluate and advance the \textit{generalizability} of manipulation policies across a spectrum of tasks, environments, and difficulty levels. To alleviate the shortage of realistic data, the challenge integrates large-scale synthetic data generation with standardized real-world robotic evaluation. Participants are encouraged to leverage synthesized state-action trials to improve general-purpose policy learning, while final assessments are conducted exclusively on unseen real-world manipulation environments. Baseline implementations, including Transformer-, Diffusion-, Vision-Language-Action, and World-Action-Model–based policies, are provided to ensure reproducibility and comparability. By coupling scalable simulation-based training with rigorous real-world validation, \textbf{RoboSynChallenge} aims to foster the development of broadly capable, data-efficient, and adaptable manipulation systems, thereby paving the way toward truly general robotic intelligence.
\end{abstract}

%\vspace{-0.15in}
% \paragraph{Keywords} Embodied AI, Sim2Real Transfer, Synthetic Data, Dexterous Manipulation

\section{Competition description}
% %\vspace{-0.1in}
% \subsection{Background and impact}
% %\vspace{-0.05in}
% Provide background on the problem approached by the competition and fields of research involved. A good outcome of a competition is to learn something new by answering a scientific question or making a significant technical advance.  Describe the scope and indicate the anticipated impact of the proposed competition (e.g., economic, humanitarian, societal).
\textbf{Background.} Developing scalable and precise robotic manipulation policies represents a critical step toward realizing the vision of \textit{Embodied Artificial Intelligence (EAI)}~\cite{Smith2012DualArm}. As a technology with broad societal and industrial impact, scalable robotic manipulation holds the potential to reduce operational costs and enhance productivity across manufacturing, household, and healthcare domains~\cite{kroemer2021review}. 

Data-driven learning methods have emerged as a principled approach for developing general-purpose robotic agents that can operate flexibly across diverse tasks and environments. In recent years, \textit{generalist policy models} have been advanced through the use of Transformers~\cite{Zhao2023ACT}, diffusion models~\cite{Chi2023DiffusionPolicy}, Vision-Language-Action (VLA) frameworks~\cite{ma2024survey}, and World Action Models (WAMs)~\cite{zhu2025unified,bi2025motus,li2026causal,ye2026world}. 
These architectures enable end-to-end learning pipelines that map multimodal observations and language instructions directly to continuous robot actions~\cite{zheng2025survey}, paving the way for more autonomous, instruction-following control systems.

As research progresses toward developing generalist manipulation policies, a central question arises: how can we fairly and comprehensively evaluate and benchmark such general-purpose robotic systems? Existing benchmarks are typically categorized as either simulation-based~\cite{james2020rlbench,mees2022calvin,liu2023libero,mu2024robotwin} or real-world~\cite{yakefu2025robochallenge,jangir2026robotarena,atreya2025roboarena,chen2026manipulationnet}.
However, unlike vision or language models, which can draw upon vast internet-scale datasets, real-world robotic benchmarks rely on data obtained through physical interaction, making data collection costly, slow, and tied to specific hardware setups~\cite{ma2024survey}.
While simulation facilitates scalable data generation~\cite{mandlekar2023mimicgen,deng2025graspvla,Liu2025DexScale,zhao2026simreal}, mismatches in dynamics, kinematics, and sensing inevitably yield a simulation-to-reality (Sim2Real) gap, resulting in degraded real-world performance~\cite{mu2024robotwin,Din2025VisionLA,Nasiriany2024RoboCasa}. Bridging this Sim2Real gap thus remains a fundamental challenge in embodied intelligence research.
Moreover, most existing benchmarks rely on fixed datasets, which assume static and controlled environments. This design misaligns with the objective of building generalist policies that must generalize to unseen objects and diverse environments.

% %\vspace{-0.1in}
\begin{figure}[htbp]
    \centering
    \includegraphics[width=1\linewidth]{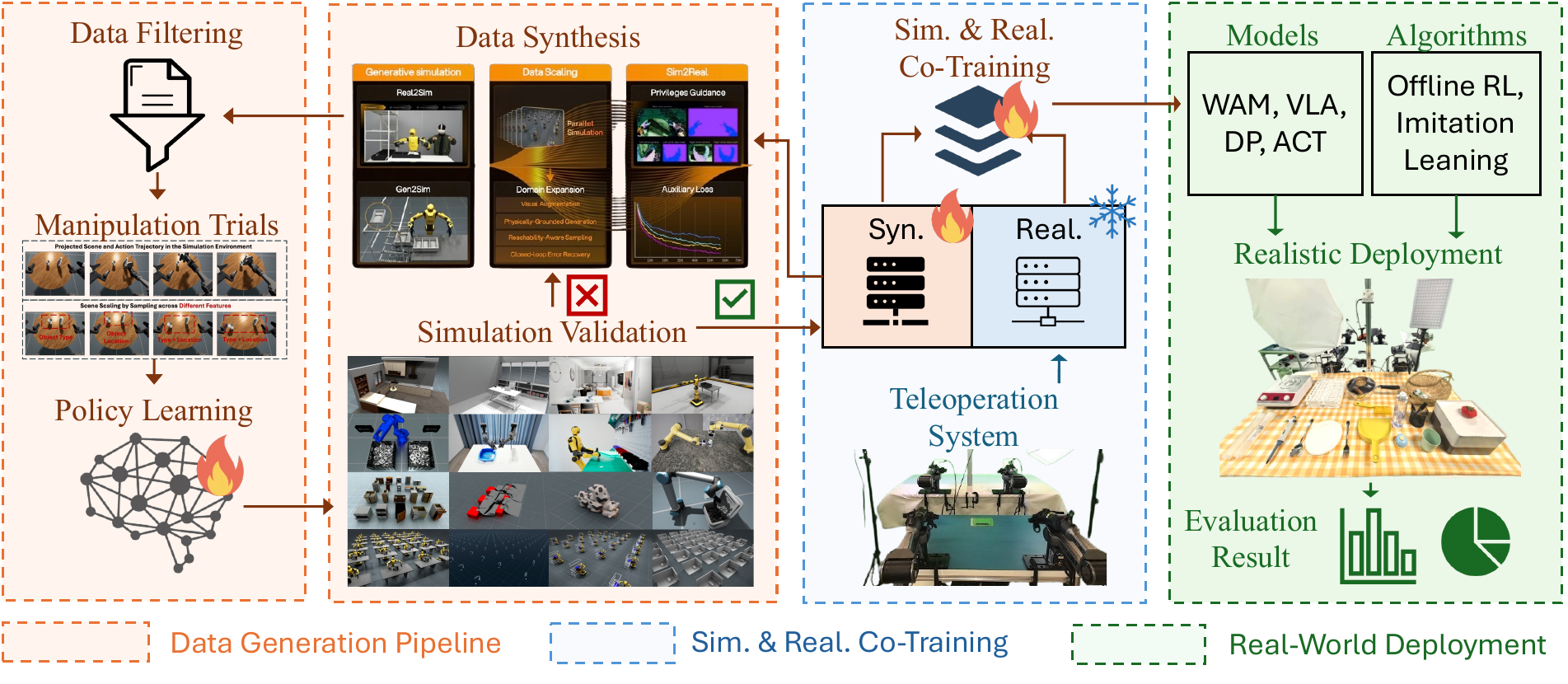}
    \caption{The pipeline of RoboSynChallenge, which provides a comprehensive data generation framework for synthesizing manipulation trials within a simulated environment (in \textcolor{orange}{orange} background), thereby expanding the training datasets. These synthetic data, together with a smaller amount of static real-world data collected through teleoperation (in \textcolor{blue}{blue} background), can be used for simulation-to-real co-training. This approach helps mitigate the limitations of scarce real data when deploying trained manipulation policies in the real world (in \textcolor{green}{green} background).}\label{fig:robosynchallenge}
    %\vspace{-0.15in}
\end{figure}

\textbf{Impact.} In this proposal, we push the frontier a step further by introducing \modelname~, a competition designed to quantify {\it how effectively simulated data can improve the generalizability of manipulation policies when real-world data is scarce}.
To achieve this goal, \modelname~ provides a comprehensive data-generation and evaluation pipeline that integrates scalable simulation environments with real-world physical setups for performance validation. The simulation platform enables practitioners to flexibly configure and scale dataset generation across diverse conditions, while the real-world evaluation protocol encompasses multiple levels of difficulty. Together, these components facilitate a rigorous and systematic analysis of Sim2Real transferability.

The \modelname~ is envisioned as an open and scalable benchmark to advance research on generalizable robot learning across simulated and real-world domains. The anticipated impact of \modelname~ is multifaceted as follows:

% \begin{itemize}
    % \item 
    \textbf{1) Standardized Benchmark.} {\modelname~} seeks to establish a standardized real-world benchmark that spans the entire dexterous manipulation policy pipeline, from large-scale data generation and policy learning with synthesized data, to rigorous evaluation under physical environments.

    \textbf{2) Real-world Reproducibility.} {\modelname~} strengthens the reproducibility of robotic control policies through standardized data synthesis protocols and evaluation frameworks, ensuring consistent benchmarking and fair comparison across real-world environments.

    \textbf{3) Opensource Tools and Baselines.} {\modelname~} provides publicly available baseline models, data synthesis, and training pipelines, fostering transparent evaluation, reproducible experimentation, and collaborative progress within the robotics research community.

    % \item 
    \textbf{4) Toward Generalizable Manipulation.} {\modelname~} provides a foundational platform toward generalizable manipulation policies by coupling large-scale synthetic data generation with standardized real-world evaluation, enabling systematic study of scalable policy generalization.
    % This integration not only facilitates the development of data-efficient and transferable policy learning methods but also provides a scalable pathway toward building unified manipulation agents capable of adapting across tasks, environments, and embodiments.
    
    % \item 
    % \textbf{5) Cross-Disciplinary Research.} {\modelname~} promotes collaboration among researchers in computer vision, robotics, and machine learning, fostering cross-disciplinary innovation and engagement across academia, industry, and open-source communities.
% \end{itemize}
% Justify the relevance of the problem to the NeurIPS community and estimate the number of participants in your proposed competition: will your competition be of interest to a large audience or limited to a small number of domain experts? 
% Describe typical real-life scenarios and/or delivery vehicles for the competition. For instance, what is the application setting, will you use a virtual or a game environment, and what situation(s)/context(s) will participants/players/agents be facing?
%\vspace{-0.15in}
\subsection{Novelty}
%\vspace{-0.05in}
% Indicate whether this is an entirely new competition, or part of a series, eventually re-using old data. If this is an updated version of a previously accepted competition (either at NeurIPS or another venue), extensively comment on the updates you introduced and provide justification if the competition is substantially similar to previous iterations. If you are aware of similar previous competitions, describe the key differences with your proposal.
As a new competition, \modelname~ introduces substantial differences compared to previous and ongoing challenges. Our key innovations are as follows:

% \begin{itemize}
\noindent {\bf 1) Benchmarking Sim2Real Transferability.} \modelname~ aims to fill a critical gap in the field by providing the first standardized benchmark for Sim2Real transferability. It systematically evaluates how well simulated data and policies support scaling to complex, real-world environments. Beyond measuring raw performance, \modelname~ assesses robustness, adaptability, and generalization under real-world noise and domain shifts.
This approach fundamentally differs from prior Sim2Sim or Real2Real benchmarks that operate within a single modality. It represents a significant step toward unified Sim2Real evaluation and understanding the limits of simulation-based learning.

\noindent {\bf 2) Generative Data Streaming.} Unlike prior benchmarks that rely on a fixed, human-crafted dataset, \modelname~ integrates an automated data generation pipeline~\cite{EmbodiChain}. The system procedurally synthesizes simulation environments and automatically generates motion trails in a closed loop. 
Moreover, to bridge the Sim2Real gap, practitioners can use the provided realistic data as input to style transfer models~\cite{ali2025world}, video generation models~\cite{wan2025wan}, or scene augmentation techniques~\cite{xue2025demogen,mandlekar2023mimicgen} to synthesize diverse motion trajectories with high visual fidelity.
This enables large-scale, diverse, and efficient data acquisition without manual curation.
% , thereby substantially increasing functional and physical diversity across training scenarios.

\noindent {\bf 3) Real-World Evaluation.}
% To ensure transfer fidelity, 
\modelname~ integrates real-world robotic evaluation as part of its official leaderboard. Submitted policies are deployed and tested on standardized physical robot platforms using the same task definitions and evaluation metrics as in simulation. This unified testbed directly measures Sim2Real performance and provides a transparent benchmark for real-world robustness, enabling participants to assess both efficiency and adaptability under tangible constraints.

% \end{itemize}

Table~\ref{table:benchmark-summary} highlights the distinctions between our approach and existing benchmarks. Unlike previous benchmarks that were limited to single-arm settings, simulations, or real-world datasets alone, \modelname~ unifies bimanual manipulation across both simulated and realistic environments. Crucially, it leverages simulated data in a generative manner to augment real-world interactions, enabling richer policy learning and improved generalization. This hybrid paradigm bridges the data efficiency of simulation with the robustness of real-world performance, extending the scope of dexterous manipulation beyond existing competitive and benchmark platforms.

%\vspace{-0.05in}
\begin{table}[htbp]
\caption{The comparison to other competition and benchmarks for dexterous manipulation.}\label{table:benchmark-summary}
\resizebox{1\textwidth}{!}{
    \begin{tabular}{ccccccccc}
    \toprule
    Competition & \begin{tabular}[c]{@{}c@{}}Rigid\\ Objects\end{tabular} & \begin{tabular}[c]{@{}c@{}}Articulated \\ Objects\end{tabular} & \begin{tabular}[c]{@{}c@{}}Assembling \\ Objects\end{tabular} & \begin{tabular}[c]{@{}c@{}}Tool \\ Using\end{tabular} & \begin{tabular}[c]{@{}c@{}}Manip.\\ Setting\end{tabular} & Dataset  &\begin{tabular}[c]{@{}c@{}}Training \\ Env.\end{tabular} & \begin{tabular}[c]{@{}c@{}}Evaluation \\ Env.\end{tabular}\\ \hline
    RLBench~\cite{james2020rlbench} & \cmark & \cmark & \xmark & \xmark & Single-Arm & Offline & Simulated & Simulated \\
    CALVIN~\cite{mees2022calvin} & \cmark & \cmark & \xmark & \xmark & Single-Arm & Offline & Simulated & Simulated \\
    LIBERO~\cite{liu2023libero} & \cmark & \cmark & \xmark & \xmark & Single-Arm & Offline & Simulated & Simulated \\
    RoboTwin~\cite{chen2025benchmarking}  & \cmark & \cmark & \xmark & \cmark & Both & Offline & Simulated & Simulated \\
    WBCD~\footnote{https://wbcdcompetition.github.io/2025/index.html}  & \cmark & \cmark & \cmark & \xmark & Bimanual & None & Realistic & Realistic \\
    RoboChallenge~\cite{yakefu2025robochallenge}  & \cmark & \cmark & \cmark & \cmark & Both & Offline & Realistic & Realistic \\
    RobotArena $\infty$~\cite{jangir2026robotarena}& \cmark & \cmark & \xmark & \xmark & Single-Ar & None & Realistic & Realistic\\
    RoboArena~\cite{atreya2025roboarena} & \cmark & \cmark & \xmark & \cmark & Single-Arm & Offline & Realistic & Simulated \\
    ManipulationNet~\cite{chen2026manipulationnet} & \cmark & \xmark & \cmark & \xmark & Single-Arm & Offline & Realistic & Realistic \\
    ManipArena~\footnote{https://maniparena.x2robot.com/} & \cmark & \cmark & \xmark & \xmark & Bimanual & Offline & Realistic & Realistic \\ 
    \modelname~ & \cmark & \cmark & \cmark & \cmark & Bimanual & Generative & Sim. \& Real. & Realistic \\ 
    \bottomrule
    \end{tabular}
}
% %\vspace{-0.1in}
\end{table}

%\vspace{-0.15in}
\subsection{Data}
% %\vspace{-0.1in}
% If your competition evaluates submissions based on the analysis of data, please provide detailed information about the availability of the evaluation data and their annotations, as well as permissions or licenses to use such data.

% If new data were collected or generated for the purpose of the competition, provide details on the procedure, including permissions to collect such data obtained by an ethics committee, if human subjects are involved, and describe steps taken to ensure the anonymity of those subjects, in accordance with the \href{https://neurips.cc/public/EthicsGuidelines}{NeurIPS Code of Ethics}. In this case, it must be clear in the document that the data will be ready and approved for use prior to the official launch of the competition. 

% \textbf{Please justify that:} 1) you have access to large enough data sets to make the competition interesting and draw conclusive and statistically significant results; 2) the data will be made freely available for the competition; 3) the ground truth has not been previously published and has been kept confidential; 4) potential feature leakage is addressed.

% Please also justify and confirm that the data complies with all data-related concerns of the \href{https://neurips.cc/public/EthicsGuidelines}{NeurIPS Code of Ethics}, including privacy, consent, deprecation, copyright and fair use, representativeness and involvement of human subjects or participants.

The dataset employed in {\modelname~} is primarily generated specifically for this competition using the \textit{EmbodiChain}~\cite{EmbodiChain} generative simulation framework. It comprises large-scale embodied interaction trajectories, multi-modal sensory streams (including RGB-D data, proprioceptive signals, and physics-based scene metadata), and structured annotations detailing task semantics, contact events, and success outcomes. 
To moderate the difficulty of Sim2Real transfer, {\modelname~} also includes a smaller subset of data collected through teleoperation to provide real-world correspondence for a limited set of scenarios. The detailed generation process is described as follows. 

All robot data were produced within robot manipulation environments using a simulation environment, generative methods, and teleoperation. No personally identifiable or human data are included.

% \textbf{Data Generation Pipeline.} \modelname~ builds on the open-source framework \textit{EmbodiChain}~\cite{EmbodiChain}, which integrates physics-based simulation, large-scale data expansion, and Sim2Real generalization into a unified process.
% % that together enable scalable, diverse, and physically grounded embodied data generation. 
% It begins with a {\it generative simulation stage} that synthesizes realistic, simulation-ready environments using generative modeling and physical optimization. This is followed by {\it domain expansion for data scaling}, where robot interaction trajectories are automatically generated, diversified, and refined through adaptive error recovery to enhance robustness and coverage. Finally, an {\it online data streaming mechanism} supports continuous Sim2Real transfer by feeding richly varied simulation experiences into the learning process, incorporating real-time visual perturbations and high-throughput data augmentation.
% Appendix~\ref{subsec:embodichain} shows the details of data generation.
% This unified design ensures both diversity and realism in large-scale embodied datasets, providing a strong foundation for robust manipulation policy learning.

\textbf{Data Collection Protocol.} To facilitate the generalizability of manipulation policy in the RoboSynChallenge, we designed protocols for collecting both simulated and real-world datasets.

1) {\it Real-World Data Collection Protocol.} To construct the real-world dataset, each manipulation task was conducted under five distinct experimental conditions, each further evaluated across four positional variations and three orientation settings, resulting in a total of 60 samples per task. The experimental conditions are designed to introduce variability in background texture, illumination, and scene complexity, thereby enabling a rigorous assessment of generalization in real-world manipulation scenarios. The detailed configuration of the experimental conditions is summarized in Table~\ref{tab:real_world_protocol} in Appendix~\ref{subsec:real-data-collection}. Each condition was combined with four positional variations and three orientation settings, ensuring comprehensive coverage of spatial and visual factors across all collected samples.

2) {\it Synthesized Data Collection Protocol.} \modelname~integrates Embodichain~\cite{EmbodiChain} to scale up the simulated state-action trails. Appendix~\ref{subsec:embodichain} shows the details of data generation.
% was generated with extensive randomization applied across environmental, object, and camera parameters to emulate realistic variability and improve model generalization. 
Each simulation instance involved systematic perturbations in lighting, object attributes, table properties, background color, and robot configuration. In addition, both intrinsic and extrinsic camera parameters were randomized to simulate different viewpoints and imaging conditions. The randomization schema is summarized in Table~\ref{tab:sim_data_protocol} in the Appendix.
This procedure introduces controlled, multi-level variability, including lighting, materials, geometry, and sensing, which ensures the simulated dataset captures diverse visual and physical conditions for robust policy training and evaluation.
% During implementation, 
We randomly sampled variations to generate 1,000 manipulation trials for each task. 
The complete data collection pipeline has been released as open-source, {\it allowing practitioners to flexibly modify and extend the framework according to their specific research needs.}

In addition, since we provide realistic data, practitioners can leverage it to guide video generation models~\cite{wan2025wan} through in-context learning or fine-tuning. The generated data can be further augmented or refined using scene augmentation techniques (e.g.,~\cite{xue2025demogen,mandlekar2023mimicgen}) applied in either simulated or real-world settings. \modelname~ offers flexible integration with such pipelines, facilitating large-scale synthesized data collection.

% \textbf{Dataset Statistics.} The dataset comprises more than xxx procedurally generated scenes and over xxx simulated or real-world collected trajectories, encompassing diverse tasks, spatial, background... textures.

\textbf{Confidential Ground Truth and Leakage Prevention.} Our evaluation protocol measures the model’s real-world performance using diverse metrics (Section~\ref{sec:metrics}). This evaluation does not require access to the testing data or any ground-truth labels for verification. Instead, it assesses outcomes directly in the real world. Moreover, the training data consist only of successful trajectories collected in the learning environment. 
To quantify generalizability, the evaluation employs a held-out testing protocol in which the test environments are out-of-distribution relative to the training observations, ensuring minimal overlap and reducing the risk of dataset leakage.
% \end{enumerate}

% \textbf{Data Availability and Licensing.}
All training and evaluation datasets will be made freely available to registered competition participants through online platforms (e.g., Hugging Face). The dataset is released under a \textbf{Creative Commons Attribution–NonCommercial–ShareAlike 4.0 (CC BY-NC-SA 4.0)} license, permitting research use while preventing commercial redistribution. Each release will include {\it metadata, documentation, and generation code} (for simulated data), in accordance with the \href{https://neurips.cc/public/EthicsGuidelines}{NeurIPS Code of Ethics}.

%\vspace{-0.15in}
\subsection{Tasks and application scenarios}
% %\vspace{-0.1in}
% Describe the competition's task(s) and explain to which specific real-world scenario(s) they correspond. Put particular emphasis on how the competition relates to a real problem faced in industry or academia. If it cannot be cast in those terms, provide a detailed hypothetical application scenario and focus on its relevance to NeurIPS.

\begin{figure}[htbp]
    \centering
    % %\vspace{-0.4in}
    \includegraphics[width=\linewidth]{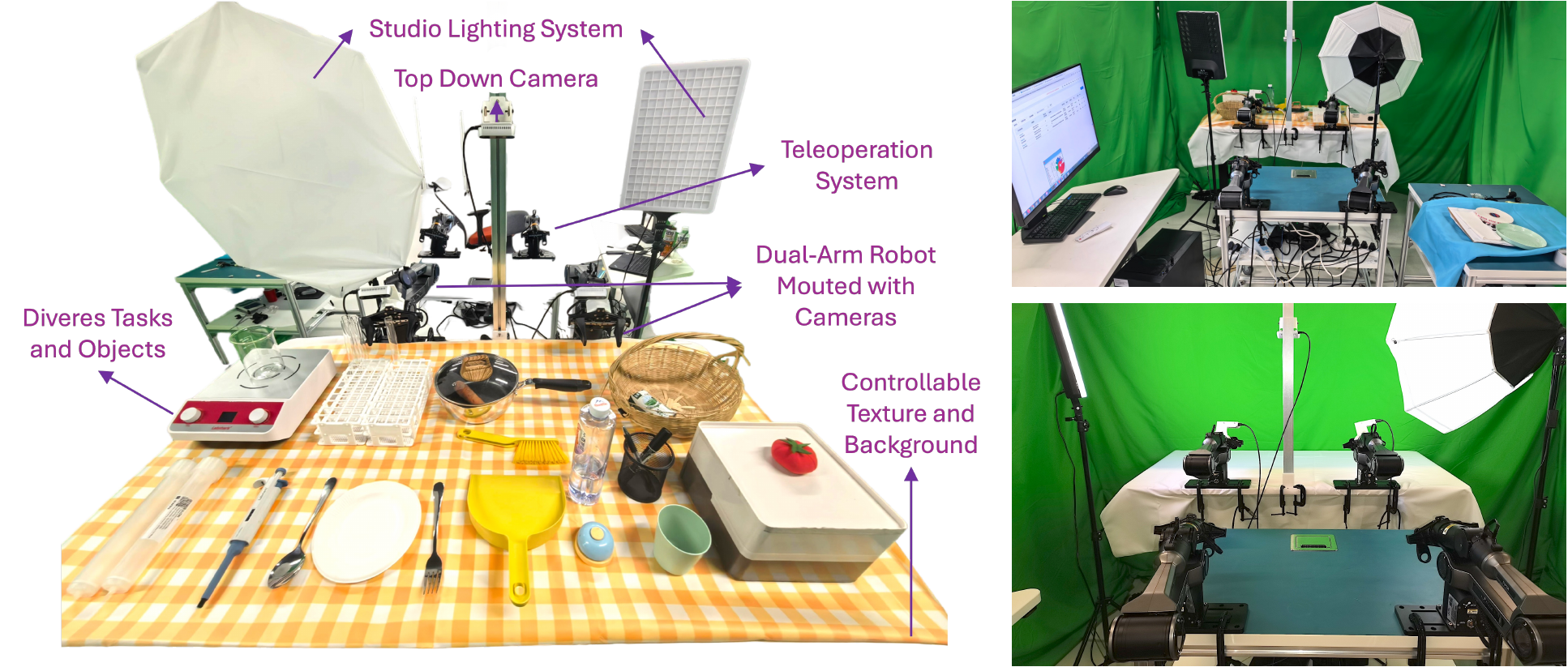}
    %\vspace{-0.3in}
    \caption{Real-world manipulation setup: The physical platform consists of a dual-arm robot, cameras, lighting, and various objects, enabling evaluation of all tasks included in the benchmark. To facilitate multiple rounds of real-world testing in a reliable and efficient manner, we have prepared {\it three sets of backup workstations}, each configured identically to the primary setup (Figure~\ref{fig:backup-hardware}).}\label{fig:real-world}
    %\vspace{-0.2in}
\end{figure}

% Our work focuses on evaluating the performance of dexterous manipulation with bimanual robots across different real-world environments. 

\textbf{Hardware.} As Figure~\ref{fig:real-world} shows, the evaluation of manipulation policy relies on a dual-arm platform built from two AgileX Piper manipulators. Each Piper is a 6-DoF research arm featuring open interfaces, designed for embodied manipulation and rapid system integration. This platform has been widely adopted in both prior competitions (e.g., RoboTwin~\cite{mu2024robotwin}, WBCD 2025~\footnote{https://wbcdcompetition.github.io/2025/index.html}, and WBCD 2026~\footnote{https://wbcdcompetition.github.io}) and recent research works~\cite{bi2025motus,wei2025cyclemanip,zhao2026simreal,wang2026evaaligningvideoworld}.

\begin{figure}
    \centering
    \includegraphics[width=0.9\linewidth]{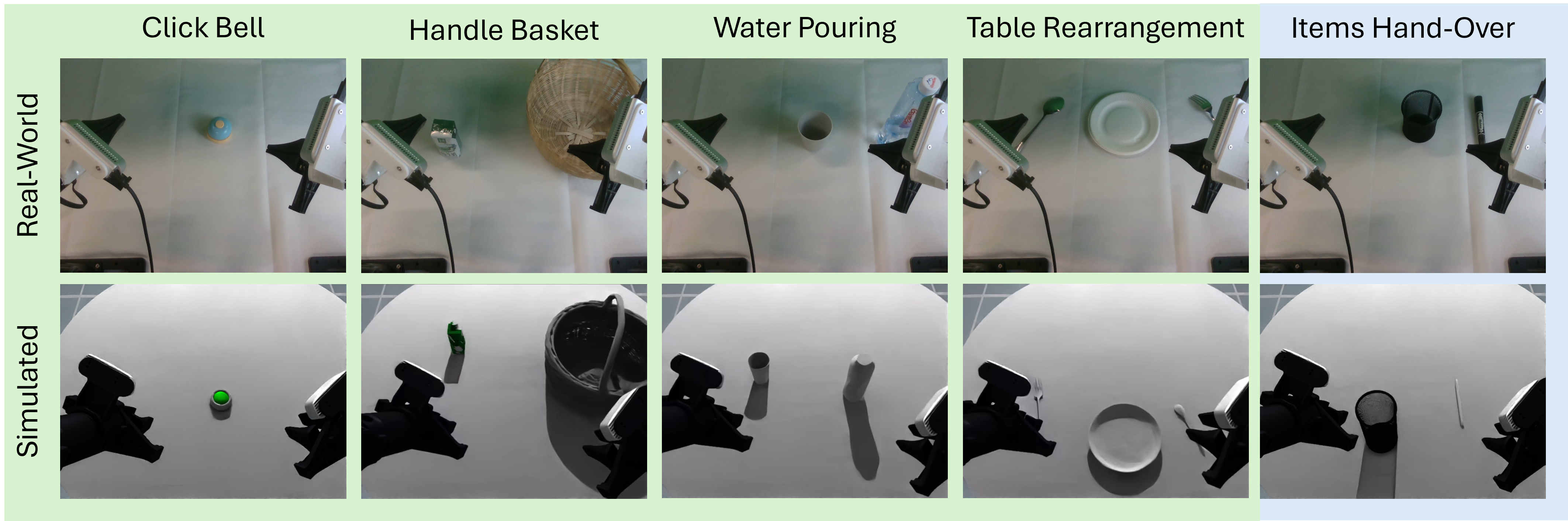}
    \includegraphics[width=0.9\linewidth]{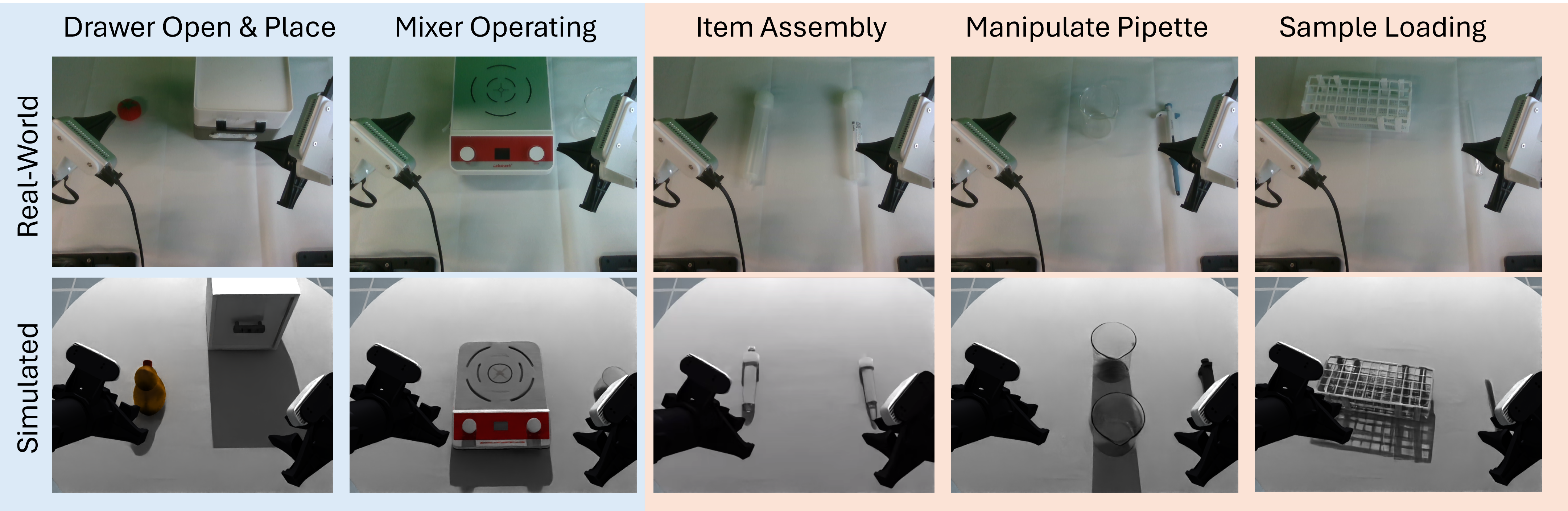}
    \caption{Visualizations of the realistic environments (upper row) and their corresponding simulated environments (lower row) across the three task categories: entry-level (green background), mid-level (blue background), and high-level (red background). }\label{fig:task-vis}
    %\vspace{-0.2in}
\end{figure}

\textbf{Manipulation tasks.} Under the above hardware system, our design tasks are diverse and involve handling rigid objects, articulated objects, deformable objects, and tools.
All tasks are performed based solely on visual observations and the robot’s proprioceptive feedback.
An overview of the real-world setup is shown in Figure~\ref{fig:real-world}. The workspace consists of an adjustable-height table with replaceable tablecloth textures and configurable lighting. 
We categorize the tasks 
% into entry-level, mid-level, and difficult-level 
as follows:

{\it 1) Entry-level tasks} primarily include short-horizon, low-contact-complexity routines with clear affordances. As shown in Figure~\ref{fig:task-vis} (green background) and Table~\ref{table:tasks}, they cover table rearrangement, click-bell, water pouring, and handle basket. These tasks emphasize stable perception, reliable grasp execution, and consistent motion trajectories under minimal environmental uncertainty. The focus lies in ensuring repeatable performance of simple action primitives rather than complex planning or intricate contact manipulation. 
% To enable clearer benchmarking, we report decomposed action sequences, associated contact phases, and explicit success metrics for each entry-level task.

{\it 2) Mid-level tasks} consist of compound, sequential interactions requiring moderate coordination, adaptive control, and spatial reasoning. As described in Figure~\ref{fig:task-vis} (blue background) and Table~\ref{table:tasks}, these include items hand-over, drawer open-and-place, and mixer operating. In contrast to entry-level routines, mid-level tasks introduce variable contact conditions, multi-object dependencies, and partial planning horizons. Evaluation focuses on assessing smooth transition across motion phases, consistent force regulation, and robustness to mild external perturbations.

{\it 3) High-level tasks} comprise fine-grained manipulation and domain-specific operations demanding high precision, multi-stage sequencing, and dynamic adaptation. As outlined in Figure~\ref{fig:task-vis} (red background) and Table~\ref{table:tasks}, they include item assembly, pipette manipulation, and sample loading. These tasks require complex reasoning over ordered procedural steps, precise end-effector control, and environmental feedback integration. Benchmarking evaluates success through accuracy of placement or alignment, procedural consistency, and recovery capability under minor execution deviations.

Across these three levels, \modelname~provides in-distribution testing (with respect to real-world training data) for model development and out-of-distribution testing, acting as a held-out benchmark to evaluate generalization quality throughout the competition.

% Justify that the problem posed is scientifically or technically challenging but not impossible to solve. If data are used, consider illustrating the same scientific problem using several data sets from other application domains.

% Additionally, please justify and confirm that the task and application scenarios comply with the \href{https://neurips.cc/public/EthicsGuidelines}{NeurIPS Code of Ethics}, including considerations regarding societal impact, potential harmful consequences, and the measures proposed to mitigate such consequences.

%\vspace{-0.15in}
\subsection{Metrics}\label{sec:metrics}
% %\vspace{-0.1in}
In our experiments, we present the evaluation protocol and metrics as follows.
% For quantitative evaluations, select one or more scoring metrics and justify that they effectively assess the efficacy of solving the problem at hand and, if relevant, that they align with real-world task performance. If no metrics are used, explain why and how the evaluation will be carried out. Explain how error bars will be computed and/or how the significance in performance differences between participants will be evaluated. If submissions must be reproduced by the organizers, the computing infrastructure used for reproduction must be described in full detail.

\textbf{Evaluation Protocol.} We evaluate dexterous manipulation performance across multiple tasks under the following real-world variations:
1) \textit{Background variation}: three table textures (wood, blue fabric, yellow grid).
2) \textit{Lighting variation}: three distinct light positions with varying illumination colors.
3) \textit{Object variation}: both \textit{seen} and \textit{unseen} object instances within the same task category, evaluating generalization to novel appearances.
4) \textit{Distractor presence}: task-irrelevant objects placed in the scene at three difficulty levels (2, 4, and 8 distractors), sampled from a separate object set distinct from task objects.
5) \textit{Spatial generalization}: evaluation on unseen positions within a predefined $3 \times 3$ grid.
% \end{itemize}
For each task, we first evaluate the policy under a canonical base configuration. We then systematically vary one factor at a time while keeping all other factors fixed. For each configuration, we evaluate the task under multiple object positions by applying small random shifts, and report the resulting success rate across these variations.

{\bf Evaluation Metrics.} Based on the aforementioned evaluation protocol, we evaluate the performance of a dexterous policy according to the following methods: {\it 1) Success Rate (SR)} (in percentage) measures how often a dexterous manipulation policy completes a task according to predefined success criteria (see Table~\ref{table:tasks}). 
2) {\it Inference Time} measures how quickly the policy produces actionable commands during inference; lighter models typically run faster. All models are deployed on the same machine with an NVIDIA A800 GPU to ensure fair comparison.
3) {\it Action Steps} count the number of control steps the model takes to finish a task, such that a higher count can indicate added latency due to error detection, feedback, and resolution, or reflect unresolved failures where the task remains incomplete, and the action budget is exhausted.

% You can include subjective measures provided by human judges. In that case, describe the judging criteria, which must be as orthogonal as possible, sensible, and specific. Provide details on the judging protocol, especially how to break ties between judges. Explain how judges will be recruited and, if possible, give a tentative list of judges, justifying their qualifications.  
%\vspace{-0.15in}
\subsection{Baselines, code, and material provided}
% %\vspace{-0.1in}
{\bf Baselines.} The baselines for our task include five representative policy families, covering sequence modeling, generative action prediction, vision-language-action policies, and world-action modeling:
1) {\bf Action Chunking Transformer (ACT)~\cite{Zhao2023ACT}}: A Transformer-based imitation learning policy that predicts temporally extended chunks of future actions from visual observations and proprioceptive states. By generating action chunks rather than single-step commands, ACT reduces compounding errors, improves temporal consistency, and provides a strong lightweight baseline for bimanual manipulation under limited demonstrations.
2) {\bf Diffusion Policy~\cite{Chi2023DiffusionPolicy}}: A visuomotor policy that formulates action prediction as a conditional denoising diffusion process. This baseline is well suited for multimodal manipulation behaviors, since it can represent diverse feasible action trajectories conditioned on the current observation history and iteratively refine them into smooth executable controls.
3) {$\boldsymbol{\pi_0}$~\cite{Intelligence2024Pi0}}: A VLA model that combines a flow-matching architecture with a pre-trained vision-language model (VLM). The flow-matching component enables smooth trajectory learning between observations and actions, while the VLM provides strong visual-semantic grounding.
4) {$\boldsymbol{\pi_{0.5}}$~\cite{Intelligence2025pi05}}: A VLA model that builds upon \(\pi_0\) by incorporating partial fine-tuning of the pre-trained vision-language backbone within the flow-matching framework. This allows \(\pi_{0.5}\) to adapt more effectively to downstream tasks while retaining strong generalization from pre-training. 
5) {\bf Motus~\cite{bi2025motus}}: A unified latent action world model that integrates vision, language, and action understanding through a mixture-of-transformers architecture. By leveraging optical-flow-based latent actions and large-scale multimodal data, Motus achieves strong generalization and state-of-the-art performance across both simulated and real-world manipulation tasks.
Together, these baselines establish a progressively richer comparison suite: ACT and Diffusion Policy represent widely used visuomotor imitation-learning methods, \(\pi_0\) and \(\pi_{0.5}\) evaluate recent VLA-style generalist policies, and Motus tests world-action-model-based reasoning over latent actions. For each baseline, we will provide standardized data loaders, training scripts, checkpoint formats, and evaluation wrappers so participants can reproduce results and compare new methods under the same Sim2Real protocol. The initial experimental results for the currently evaluated VLA and WAM baselines, trained using simulation-only or real-only data, are summarized in Table~\ref{table:results_final} below. Furthermore, we plan to incorporate a broader range of state-of-the-art baselines in future work, including VLA models such as {\bf RDT-1B}~\cite{liu2025rdtb} and WAMs like {\bf DreamZero}~\cite{ye2026world}.

\begin{table}[h!]
\centering
\renewcommand{\arraystretch}{1.5}
\setlength{\tabcolsep}{6pt}
\caption{Experimental Results of Baselines: Success Rate ($x/20$ or \%), Action Steps (max 1000), and Real Time (s).}
\label{table:results_final}
\small

\resizebox{1\textwidth}{!}{
\begin{tabular}{l | ccc | ccc | ccc | ccc}
\hline
\multirow{2}{*}{\textbf{Model}} & \multicolumn{3}{c|}{\textbf{Click Bell}} & \multicolumn{3}{c|}{\textbf{Items Hand-Over and Place}} & \multicolumn{3}{c|}{\textbf{Dual-Arm Water Pouring}} & \multicolumn{3}{c}{\textbf{Table Rearrangement}} \\
\cline{2-13}
& \textbf{SR} & \textbf{Steps} & \textbf{Time} & \textbf{SR} & \textbf{Steps} & \textbf{Time} & \textbf{SR} & \textbf{Steps} & \textbf{Time} & \textbf{SR} & \textbf{Steps} & \textbf{Time} \\
\hline
\textbf{pi0 (sim)}   & 8/20 & 625.30 & \textbf{63.78} & 5/20 & 834.75 & 83.60 & 6/20 & 898.60 & 89.95 & 7/20 & 788.20 & 79.07 \\
\textbf{pi0 (real)}  & 5/20 & 860.45 & 86.05 & 5/20 & 844.00 & 86.10 & 4/20 & 917.70 & 92.69 & 8/20 & 742.70 & 74.27 \\
\textbf{pi0.5 (sim)} & 10/20 & 647.55 & 65.53 & 7/20 & 791.25 & \textbf{80.00} & 7/20 & 877.45 & \textbf{87.90} & \textbf{12/20} & \textbf{612.90} & \textbf{61.31} \\
\textbf{pi0.5 (real)}& 6/20 & 791.20 & 80.70 & 5/20 & 832.90 & 84.12 & 6/20 & 872.15 & \textbf{87.22} & \textbf{12/20} & 628.80 & 63.51 \\
\textbf{Motus (sim)} & \textbf{13/20} & \textbf{463.30} & 79.09 & \textbf{10/20} & 584.70 & 97.52 & \textbf{8/20} & \textbf{667.30} & 111.00 & 4/20 & 864.25 & 143.75 \\
\textbf{Motus (real)}& \textbf{14/20} & \textbf{420.50} & 70.11 & \textbf{12/20} & \textbf{492.30} & 83.65 & 6/20 & 744.95 & 125.97 & 3/20 & 900.05 & 153.78 \\
\hline
\end{tabular}
}

%\vspace{1.2em}

% --- 第二组：4个任务 (与文字同宽) ---
\resizebox{1\textwidth}{!}{
\begin{tabular}{l | ccc | ccc | ccc | ccc}
\hline
\multirow{2}{*}{\textbf{Model}} & \multicolumn{3}{c|}{\textbf{Basket Pick-and-Place}} & \multicolumn{3}{c|}{\textbf{Drawer Open and Place}} & \multicolumn{3}{c|}{\textbf{Mixer Operating}} & \multicolumn{3}{c}{\textbf{Item Assembly}} \\
\cline{2-13}
& \textbf{SR} & \textbf{Steps} & \textbf{Time} & \textbf{SR} & \textbf{Steps} & \textbf{Time} & \textbf{SR} & \textbf{Steps} & \textbf{Time} & \textbf{SR} & \textbf{Steps} & \textbf{Time} \\
\hline
\textbf{pi0 (sim)}   & 5/20 & 835.40 & 83.56 & 6/20 & 821.40 & 82.23 & 3/20 & 897.05 & 90.09 & 0/20 & 1000.00 & 102.07 \\
\textbf{pi0 (real)}  & 6/20 & 796.70 & 80.47 & 8/20 & 757.70 & 77.29 & 1/20 & 967.55 & 98.69 & 0/20 & 1000.00 & 99.86 \\
\textbf{pi0.5 (sim)} & \textbf{10/20} & 663.25 & \textbf{66.42} & 11/20 & 664.40 & \textbf{67.08} & \textbf{4/20} & \textbf{864.50} & \textbf{87.11} & 0/20 & 1000.00 & 104.29 \\
\textbf{pi0.5 (real)}& 9/20 & 697.90 & 71.88 & \textbf{12/20} & 645.70 & \textbf{65.22} & 3/20 & 901.75 & 90.18 & 0/20 & 1000.00 & 101.02 \\
\textbf{Motus (sim)} & \textbf{10/20} & 827.95 & 132.29 & 10/20 & 593.15 & 102.70 & 2/20 & 936.25 & 154.80 & 0/20 & 1000.00 & 166.22 \\
\textbf{Motus (real)}& 9/20 & \textbf{608.55} & 101.27 & 11/20 & \textbf{546.60} & 93.96 & 0/20 & 1000.00 & 166.41 & 0/20 & 1000.00 & 167.37 \\
\hline
\end{tabular}
}

%\vspace{1.2em}

\makebox[\textwidth][c]{
\resizebox{0.8\textwidth}{!}{
\begin{tabular}{l | ccc | ccc | ccc}
\hline
\multirow{2}{*}{\textbf{Model}} & \multicolumn{3}{c|}{\textbf{Manipulate Pipette}} & \multicolumn{3}{c|}{\textbf{Sample Loading}} & \multicolumn{3}{c}{\textbf{Task Average}} \\
\cline{2-10}
& \textbf{SR} & \textbf{Steps} & \textbf{Time} & \textbf{SR} & \textbf{Steps} & \textbf{Time} & \textbf{SR} & \textbf{Steps} & \textbf{Time} \\
\hline
\textbf{pi0 (sim)}   & 2/20 & 960.65 & 96.29 & 2/20 & 946.85 & 94.73 & 22.00\% & 898.12 & 90.56 \\
\textbf{pi0 (real)}  & 0/20 & 1000.00 & 108.46 & 3/20 & 920.35 & 92.04 & 22.50\% & 881.15 & 90.20 \\
\textbf{pi0.5 (sim)} & 2/20 & 953.15 & \textbf{95.82} & \textbf{4/20} & \textbf{900.05} & \textbf{89.97} & \textbf{38.50\%} & 797.55 & \textbf{80.55} \\
\textbf{pi0.5 (real)}& \textbf{4/20} & \textbf{898.75} & \textbf{91.67} & 3/20 & 914.95 & 93.32 & 33.00\% & 821.65 & 82.35 \\
\textbf{Motus (sim)} & \textbf{4/20} & 905.35 & 149.33 & 2/20 & 945.70 & 157.82 & 31.50\% & 778.80 & 133.76 \\
\textbf{Motus (real)}& 0/20 & 1000.00 & 164.84 & 0/20 & 1000.00 & 166.85 & 27.50\% & \textbf{721.35} & 129.43 \\
\hline
\end{tabular}
}
}
\end{table}

% Indicate \textbf{how} and \textbf{when} you plan to release the ``starting kit'' for your competition. This should include code for baselines and data-loading tools to help the participants easily join the competition. For certain competitions, the material provided may include a hardware platform.

{\bf Protocol of Releasing Code and Data} We plan to release the starting kit through a dedicated GitHub repository that will contain all necessary materials for participants to easily join the competition. The repository will include:
1) Baseline code for model training and evaluation,
2) Data‑loading and preprocessing tools for both simulated and real‑world datasets, and 
3) Sample simulated data together with a data‑generation pipeline to enable participants to reproduce and extend training environments.
In addition, we will release the real‑world dataset and the complete simulated dataset (with data generator) on a public hosting platform (Hugging Face) to ensure accessibility and scalability. 
% Depending on data size and access control, we plan to use Hugging Face Datasets Hub

% The starting kit and datasets will be made available at the competition launch, with the baseline code and sample data released in a public beta approximately two weeks before the official start, allowing participants to set up their environments and provide early feedback

%\vspace{-0.15in}
\subsection{Website, tutorial, and documentation}\label{subsec:web-tutoral-docu}
% %\vspace{-0.1in}
A dedicated competition website\footnote{Competition website: \url{https://robosyn-bench.net/}} will serve as the central hub for all information related to the event. The website will comprehensively present the competition overview, detailed timeline, participation steps, and relevant submission instructions. It will feature a FAQ/Tutorial section offering step-by-step guidance on installing the simulation environment, generating datasets, and testing models in both simulated and real-world settings. To ensure open communication, participants will be able to contact the organizers directly via robosynchallenge@gmail.com, with responses and updates maintained regularly. All content, including documentation and tutorials, will be made available well in advance of the competition start date, and the website will go live within two weeks of acceptance notification. The official GitHub repository \footnote{GitHub repository: \url{https://github.com/EDEM-AI/RoboSynChallenge}} will host the codebase, baseline models, and data tools, while an accompanying white paper will describe the competition design, problem formulation, and technical foundation supporting the challenge.

% \end{itemize}
% \subsection{Support requested}

% Please indicate the kind of support you need from the conference and \textbf{keep in mind that the NeurIPS 2025 Competition Track is a \emph{in person} event.}

\bibliography{bibliography}

@inproceedings{zhao2026simreal,
title={Sim2Real {VLA}: Zero-Shot Generalization of Synthesized Skills to Realistic Manipulation},
author={Runyi Zhao and Sheng Xu and Ruixing Jin and Yueci Deng and Yunxin Tai and Kui Jia and Guiliang Liu},
booktitle={International Conference on Learning Representations, {ICLR}},
year={2026},
}

@article{Intelligence2024Pi0,
  author       = {Kevin Black and
                  Noah Brown and
                  Danny Driess and
                  Adnan Esmail and
                  Michael Equi and
                  Chelsea Finn and
                  Niccolo Fusai and
                  Lachy Groom and
                  Karol Hausman and
                  Brian Ichter and
                  Szymon Jakubczak and
                  Tim Jones and
                  Liyiming Ke and
                  Sergey Levine and
                  Adrian Li{-}Bell and
                  Mohith Mothukuri and
                  Suraj Nair and
                  Karl Pertsch and
                  Lucy Xiaoyang Shi and
                  James Tanner and
                  Quan Vuong and
                  Anna Walling and
                  Haohuan Wang and
                  Ury Zhilinsky},
  title        = {{\(\pi\)}\({}_{0}\): {A} Vision-Language-Action Flow Model for General Robot Control},
  journal      = {arXiv preprint arXiv:2410.24164},
  year         = {2024},
}

@article{Intelligence2025pi05,
  title={$\pi$${}_{0.5}$: a Vision-Language-Action Model with Open-World Generalization},
  author={Intelligence, Physical and Black, Kevin and Brown, Noah and Darpinian, James and Dhabalia, Karan and Driess, Danny and Esmail, Adnan and Equi, Michael and Finn, Chelsea and Fusai, Niccolo and others},
  journal={arXiv preprint arXiv:2504.16054},
  year={2025}
}

@article{Din2025VisionLA,
  title={Vision Language Action Models in Robotic Manipulation: A Systematic Review},
  author={Muhayy ud Din and Waseem Akram and Lyes Saad Saoud and Jan Rosell and Irfan Hussain},
  journal={ArXiv},
  year={2025},
  volume={abs/2507.10672},
}

@article{chen2025benchmarking,
  title={Benchmarking Generalizable Bimanual Manipulation: RoboTwin Dual-Arm Collaboration Challenge at CVPR 2025 MEIS Workshop},
  author={Chen, Tianxing and Wang, Kaixuan and Yang, Zhaohui and Zhang, Yuhao and Chen, Zanxin and Chen, Baijun and Dong, Wanxi and Liu, Ziyuan and Chen, Dong and Yang, Tianshuo and others},
  journal={arXiv preprint arXiv:2506.23351},
  year={2025}
}

@inproceedings{
liu2025rdtb,
title={{RDT}-1B: a Diffusion Foundation Model for Bimanual Manipulation},
author={Songming Liu and Lingxuan Wu and Bangguo Li and Hengkai Tan and Huayu Chen and Zhengyi Wang and Ke Xu and Hang Su and Jun Zhu},
booktitle={International Conference on Learning Representations, ICLR},
year={2025},
}

@article{liu2023libero,
  title={LIBERO: Benchmarking Knowledge Transfer for Lifelong Robot Learning},
  author={Liu, Bo and Zhu, Yifeng and Gao, Chongkai and Feng, Yihao and Liu, Qiang and Zhu, Yuke and Stone, Peter},
  journal={arXiv preprint arXiv:2306.03310},
  year={2023}
}

@inproceedings{atreya2025roboarena,
  title = {RoboArena: Distributed Real-World Evaluation of Generalist Robot Policies},
  author = {Atreya, Pranav and Pertsch, Karl and Lee, Tony and Kim, Moo Jin and Jain, Arhan and Kuramshin, Artur and Eppner, Clemens and Neary, Cyrus and Hu, Edward and Ramos, Fabio and others},
  booktitle = {Proceedings of the Conference on Robot Learning (CoRL 2025)},
  year = {2025}
}

@article{mees2022calvin,
author = {Oier Mees and Lukas Hermann and Erick Rosete-Beas and Wolfram Burgard},
title = {CALVIN: A Benchmark for Language-Conditioned Policy Learning for Long-Horizon Robot Manipulation Tasks},
journal={IEEE Robotics and Automation Letters (RA-L)},
volume={7},
number={3},
pages={7327-7334},
year={2022}
}

@article{james2020rlbench,
  title={Rlbench: The robot learning benchmark \& learning environment},
  author={James, Stephen and Ma, Zicong and Arrojo, David Rovick and Davison, Andrew J},
  journal={IEEE Robotics and Automation Letters},
  volume={5},
  number={2},
  pages={3019--3026},
  year={2020},
  publisher={IEEE}
}

@inproceedings{mandlekar2023mimicgen,
    title={MimicGen: A Data Generation System for Scalable Robot Learning using Human Demonstrations},
    author={Mandlekar, Ajay and Nasiriany, Soroush and Wen, Bowen and Akinola, Iretiayo and Narang, Yashraj and Fan, Linxi and Zhu, Yuke and Fox, Dieter},
    booktitle={Annual Conference on Robot Learning, CoRL},
    year={2023}
}

@article{Nasiriany2024RoboCasa,
  title={RoboCasa: Large-Scale Simulation of Everyday Tasks for Generalist Robots},
  author={Nasiriany, Soroush and Maddukuri, Abhiram and Zhang, Lance and Parikh, Adeet and Lo, Aaron and Joshi, Abhishek and Mandlekar, Ajay and Zhu, Yuke},
  journal={arXiv preprint arXiv:2406.02523},
  year={2024}
}

@inproceedings{Liu2025DexScale,
  author       = {Guiliang Liu and
                  Yueci Deng and
                  Runyi Zhao and
                  Huayi Zhou and
                  Jian Chen and
                  Jietao Chen and
                  Ruiyan Xu and
                  Yunxin Tai and
                  Kui Jia},
  title        = {DexScale: Automating Data Scaling for Sim2Real Generalizable Robot Control},
  booktitle    = {International Conference on Machine Learning, {ICML}},
  year         = {2025},
}

@article{mu2024robotwin,
  title={Robotwin: Dual-arm robot benchmark with generative digital twins (early version)},
  author={Mu, Yao and Chen, Tianxing and Peng, Shijia and Chen, Zanxin and Gao, Zeyu and Zou, Yude and Lin, Lunkai and Xie, Zhiqiang and Luo, Ping},
  journal={arXiv preprint arXiv:2409.02920},
  year={2024}
}

@inproceedings{Chi2023DiffusionPolicy,
  author       = {Cheng Chi and
                  Siyuan Feng and
                  Yilun Du and
                  Zhenjia Xu and
                  Eric Cousineau and
                  Benjamin Burchfiel and
                  Shuran Song},
  title        = {Diffusion Policy: Visuomotor Policy Learning via Action Diffusion},
  booktitle    = {Robotics: Science and Systems, {RSS}},
  year         = {2023},
}

@inproceedings{Zhao2023ACT,
  author       = {Tony Z. Zhao and
                  Vikash Kumar and
                  Sergey Levine and
                  Chelsea Finn},
  title        = {Learning Fine-Grained Bimanual Manipulation with Low-Cost Hardware},
  booktitle    = {Robotics: Science and Systems, RSS},
  year         = {2023}
}

@article{ma2024survey,
  title={A survey on vision-language-action models for embodied ai},
  author={Ma, Yueen and Song, Zixing and Zhuang, Yuzheng and Hao, Jianye and King, Irwin},
  journal={arXiv preprint arXiv:2405.14093},
  year={2024}
}

@article{zheng2025survey,
  title={A survey of embodied learning for object-centric robotic manipulation},
  author={Zheng, Ying and Yao, Lei and Su, Yuejiao and Zhang, Yi and Wang, Yi and Zhao, Sicheng and Zhang, Yiyi and Chau, Lap-Pui},
  journal={Machine Intelligence Research},
  pages={1--39},
  year={2025},
  publisher={Springer}
}

@article{deng2025graspvla,
  title={Graspvla: a grasping foundation model pre-trained on billion-scale synthetic action data},
  author={Deng, Shengliang and Yan, Mi and Wei, Songlin and Ma, Haixin and Yang, Yuxin and Chen, Jiayi and Zhang, Zhiqi and Yang, Taoyu and Zhang, Xuheng and Zhang, Wenhao and others},
  journal={arXiv preprint arXiv:2505.03233},
  year={2025}
}

@article{Smith2012DualArm,
  author       = {Christian Smith and
                  Yiannis Karayiannidis and
                  Lazaros Nalpantidis and
                  Xavi Gratal and
                  Peng Qi and
                  Dimos V. Dimarogonas and
                  Danica Kragic},
  title        = {Dual arm manipulation - {A} survey},
  journal      = {Robotics and Autonomous Systems},
  volume       = {60},
  number       = {10},
  pages        = {1340--1353},
  year         = {2012}
}

@misc{EmbodiChain,
  author = {EmbodiChain Developers},
  title = {EmbodiChain: An end-to-end, GPU-accelerated, and modular platform for building generalized Embodied Intelligence.},
  month = {November},
  year = {2025},
  url = {https://github.com/DexForce/EmbodiChain}
}

@inproceedings{jangir2026robotarena,
title={RobotArena \${\textbackslash}infty\$: Unlimited Robot Benchmarking via Real-to-Sim Translation},
author={Yash Jangir and Yidi Zhang and Kashu Yamazaki and Chenyu Zhang and Kuan-Hsun Tu and Tsung-Wei Ke and Lei Ke and Yonatan Bisk and Katerina Fragkiadaki},
booktitle={International Conference on Learning Representations, ICLR},
year={2026}
}

@article{kroemer2021review,
  title={A review of robot learning for manipulation: Challenges, representations, and algorithms},
  author={Kroemer, Oliver and Niekum, Scott and Konidaris, George},
  journal={Journal of Machine Learning Research},
  volume={22},
  number={30},
  pages={1--82},
  year={2021}
}

@article{zhu2025unified,
  title={Unified world models: Coupling video and action diffusion for pretraining on large robotic datasets},
  author={Zhu, Chuning and Yu, Raymond and Feng, Siyuan and Burchfiel, Benjamin and Shah, Paarth and Gupta, Abhishek},
  journal={arXiv preprint arXiv:2504.02792},
  year={2025}
}

@article{bi2025motus,
  title={Motus: A unified latent action world model},
  author={Bi, Hongzhe and Tan, Hengkai and Xie, Shenghao and Wang, Zeyuan and Huang, Shuhe and Liu, Haitian and Zhao, Ruowen and Feng, Yao and Xiang, Chendong and Rong, Yinze and others},
  journal={arXiv preprint arXiv:2512.13030},
  year={2025}
}

@article{ye2026world,
  title={World action models are zero-shot policies},
  author={Ye, Seonghyeon and Ge, Yunhao and Zheng, Kaiyuan and Gao, Shenyuan and Yu, Sihyun and Kurian, George and Indupuru, Suneel and Tan, You Liang and Zhu, Chuning and Xiang, Jiannan and others},
  journal={arXiv preprint arXiv:2602.15922},
  year={2026}
}

@article{li2026causal,
  title={Causal World Modeling for Robot Control},
  author={Li, Lin and Zhang, Qihang and Luo, Yiming and Yang, Shuai and Wang, Ruilin and Han, Fei and Yu, Mingrui and Gao, Zelin and Xue, Nan and Zhu, Xing and others},
  journal={arXiv preprint arXiv:2601.21998},
  year={2026}
}

@article{yakefu2025robochallenge,
  title={RoboChallenge: Large-scale Real-robot Evaluation of Embodied Policies},
  author={Yakefu, Adina and Xie, Bin and Xu, Chongyang and Zhang, Enwen and Zhou, Erjin and Jia, Fan and Yang, Haitao and Fan, Haoqiang and Zhang, Haowei and Peng, Hongyang and others},
  journal={arXiv preprint arXiv:2510.17950},
  year={2025}
}

@article{xue2025demogen,
  title={Demogen: Synthetic demonstration generation for data-efficient visuomotor policy learning},
  author={Xue, Zhengrong and Deng, Shuying and Chen, Zhenyang and Wang, Yixuan and Yuan, Zhecheng and Xu, Huazhe},
  journal={arXiv preprint arXiv:2502.16932},
  year={2025}
}

@article{ali2025world,
  title={World simulation with video foundation models for physical ai},
  author={Ali, Arslan and Bai, Junjie and Bala, Maciej and Balaji, Yogesh and Blakeman, Aaron and Cai, Tiffany and Cao, Jiaxin and Cao, Tianshi and Cha, Elizabeth and Chao, Yu-Wei and others},
  journal={arXiv preprint arXiv:2511.00062},
  year={2025}
}

@article{wan2025wan,
  title={Wan: Open and advanced large-scale video generative models},
  author={Wan, Team and Wang, Ang and Ai, Baole and Wen, Bin and Mao, Chaojie and Xie, Chen-Wei and Chen, Di and Yu, Feiwu and Zhao, Haiming and Yang, Jianxiao and others},
  journal={arXiv preprint arXiv:2503.20314},
  year={2025}
}

@article{chen2026manipulationnet,
  title={ManipulationNet: An Infrastructure for Benchmarking Real-World Robot Manipulation with Physical Skill Challenges and Embodied Multimodal Reasoning},
  author={Chen, Yiting and Kimble, Kenneth and Adelson, Edward H and Asfour, Tamim and Chanrungmaneekul, Podshara and Chitta, Sachin and Chitambar, Yash and Chen, Ziyang and Goldberg, Ken and Kragic, Danica and others},
  journal={arXiv preprint arXiv:2603.04363},
  year={2026}
}

@article{wei2025cyclemanip,
  title={CycleManip: Enabling Cyclic Task Manipulation via Effective Historical Perception and Understanding},
  author={Wei, Yi-Lin and Liao, Haoran and Lin, Yuhao and Wang, Pengyue and Liang, Zhizhao and Liu, Guiliang and Zheng, Wei-Shi},
  journal={arXiv preprint arXiv:2512.01022},
  year={2025}
}

@article{wang2026evaaligningvideoworld,
  title={EVA: Aligning Video World Models with Executable Robot Actions via Inverse Dynamics Rewards},
  author={Ruixiang Wang and Qingming Liu and Yueci Deng and Guiliang Liu and Zhen Liu and Kui Jia},
  year={2026},
  archivePrefix={arXiv},
}
\bibliographystyle{unsrt}

\appendix

\section{Technical Details}

\subsection{Data Collection Protocol in the Real World
}\label{subsec:real-data-collection}

To evaluate real-world generalization, each manipulation task was conducted under five experimental conditions (varying background, lighting, and distractors, seen in Table~\ref{tab:real_world_protocol}), combined with four positions and three orientations. This systematic setup yields 60 diverse samples per task to capture varied spatial and visual factors.

\begin{table}[h]
\centering
%\vspace{-0.05in}
\caption{Summary of real-world data collection conditions.}\label{tab:real_world_protocol}
%\vspace{-0.05in}
\resizebox{1.0\textwidth}{!}{
\begin{tabular}{cllll}
\toprule
 \textbf{Background} & \textbf{Lighting Setting} & \textbf{Additionals} & \textbf{Description} \\ 
\hline
White & Fixed lighting & None & Standard setting with neutral background and lighting. \\ 
White & Enhanced lighting & None & Increased illumination to vary lighting conditions. \\ 
White & Fixed lighting & 2–3 distractors & Includes additional objects to test robustness. \\ 
Blue & Fixed lighting & None & Alternative background color to introduce visual contrast. \\ 
Yellow & Fixed lighting & None & Textured background used to assess visual generalization. \\ 
\bottomrule
\end{tabular}
%\vspace{-0.2in}
}
\end{table}

\subsection{Data Generation with EmbodiChain}\label{subsec:embodichain}

\textit{EmbodiChain}~\cite{EmbodiChain} is an end-to-end, GPU-accelerated, and modular platform for embodied AI research that integrates high-performance simulation, automated data pipelines, and flexible learning tools, thereby supporting rapid experimentation and effective Sim2Real transfer.
It consists of three tightly coupled stages for scalable, diverse, and physically-grounded dataset generation.

\textit{1) Generative Simulation of Learning Environment.}
To overcome the limited diversity of manually designed simulation environments, {EmbodiChain} employs a two-stage generative framework. 
First, it synthesizes simulation-ready assets via generative models followed by multi-objective optimization to ensure geometric fidelity, physical plausibility, and simulator compatibility. 
Second, these assets are composed into fully functional scenes through gradient-based layout synthesis that optimizes object placement for physical realism and robot reachability. 
This process yields physically consistent, richly annotated environments that form the foundation of large-scale embodied data generation.

\textit{2) Data Scaling via Domain Expansion.}
Building on the generated environments, EmbodiChain scales embodied data by automatically generating and expanding robot interaction trajectories to improve coverage and robustness. It promotes functional diversity through reachability-aware sampling that selects kinematically feasible robot states maximizing task-space dissimilarity (e.g., in end-effector approach direction, contact geometry, and interaction outcomes), reducing trajectory homogenization common in teleoperation and conventional planners. To further strengthen robustness, a closed-loop error recovery module detects execution failures (e.g., slippage, misaligned grasps, boundary violations) and reactively replans corrective motions, which are relabeled and reintegrated as supervision for recovery behaviors.

\textit{3) Sim2Real Generalization via Online Data Streaming.}
To facilitate scalable Sim2Real transfer, {EmbodiChain} implements an \textit{Online Data Streaming (ODS)} mechanism that continuously feeds diverse experiences from simulation into the learning loop. 
A streaming-based visual augmentation module perturbs lighting, textures, and sensor parameters on the fly, enriching perceptual diversity and mitigating overfitting to simulation-specific appearances. 
In parallel, an asynchronous shared-memory architecture enables real-time data exchange between simulation and learning processes through lock-free circular buffers, maximizing experience throughput and sample efficiency.

% \noindent Collectively, these components enable {EmbodiChain} to generate large-scale, physically consistent, and perceptually diverse datasets that support robust policy learning and realistic Sim2Real generalization.

\begin{table}[h]
\centering
\caption{Summary of simulated data randomization parameters.}
\label{tab:sim_data_protocol}
\resizebox{1.0 \textwidth}{!}{
\begin{tabular}{lll}
\hline
\textbf{Feature} & \textbf{Parameter} & \textbf{Description} \\
\hline
\multirow{3}{*}{Light} 
 & intensity\_range: [\textit{min}, \textit{max}] & Lighting intensity variation. \\
 & position\_range: [[$x_{\text{min}}$, $y_{\text{min}}$, $z_{\text{min}}$], [$x_{\text{max}}$, $y_{\text{max}}$, $z_{\text{max}}$]] & Light source position randomization. \\
 & color\_range: [[$0.6$, $0.6$, $0.6$], [$1.0$, $1.0$, $1.0$]] & Light color variation. \\
\hline
\multirow{2}{*}{Object location} 
 & init\_pos: [$x$, $y$, $z$] & Object initial position. \\
 & position\_range: [[$-x$, $-y$, $-z$], [$+x$, $+y$, $+z$]] & Offset range from initial position. \\
\hline
Object orientation & rotation\_range: [\textit{min}, \textit{max}] & Initial object rotation randomization. \\
Object material & base\_color\_range: [[$0.2$, $0.2$, $0.2$], [$1.0$, $1.0$, $1.0$]] & Object surface color variability. \\
Object size & scale: [$x$, $y$, $z$] & Uniform or axis-specific scaling. \\
\hline
\multirow{3}{*}{Table} 
 & position\_range: [[$0$, $0$, $-0.04$], [$0$, $0$, $0.04$]] & Table height variation (±4 cm). \\
 & random\_texture\_prob: $p_{\text{texture}}$ & Probability of random table texture. \\
 & base\_color\_range: [[$r_{\text{min}}$, $g_{\text{min}}$, $b_{\text{min}}$], [$r_{\text{max}}$, $g_{\text{max}}$, $b_{\text{max}}$]] & Table color range. \\
\hline
Background color & base\_color\_range: [[$0.2$, $0.2$, $0.2$], [$1.0$, $1.0$, $1.0$]] & Background plate color variation. \\
\hline
\multirow{2}{*}{Camera (intrinsics)} 
 & $f_{x,\text{range}}$ & Focal length range along x-axis. \\
 & $f_{y,\text{range}}$ & Focal length range along y-axis. \\
\hline
\multirow{2}{*}{Camera (extrinsics)} 
 & pos\_range: [[$0$, $-0.02$, $0$], [$0.02$, $0.02$, $0.01$]] & Random camera position (XYZ). \\
 & euler\_range: [[$-0.175$, $-0.175$, $-0.175$], [$0.175$, $0.175$, $0.175$]] & Camera rotation (roll–pitch–yaw). \\
\hline
\multirow{2}{*}{Robot init.} 
 & eef\_pos\_range: [[$-0.01$, $-0.01$, $-0.01$], [$0.01$, $0.01$, $0$]] & End-effector position variation. \\
 & $q_{\text{pos,range}}$: ([$j_{1,\text{min}}$,…],[$j_{1,\text{max}}$,…]) & Joint configuration randomization. \\
\hline
Distractors & Common items (e.g., bowls, cups, toys) & Distractors for scene complexity. \\
\hline
\end{tabular}
}
\end{table}

\begin{figure}[htbp]
    \centering
    \includegraphics[width=0.9\linewidth]{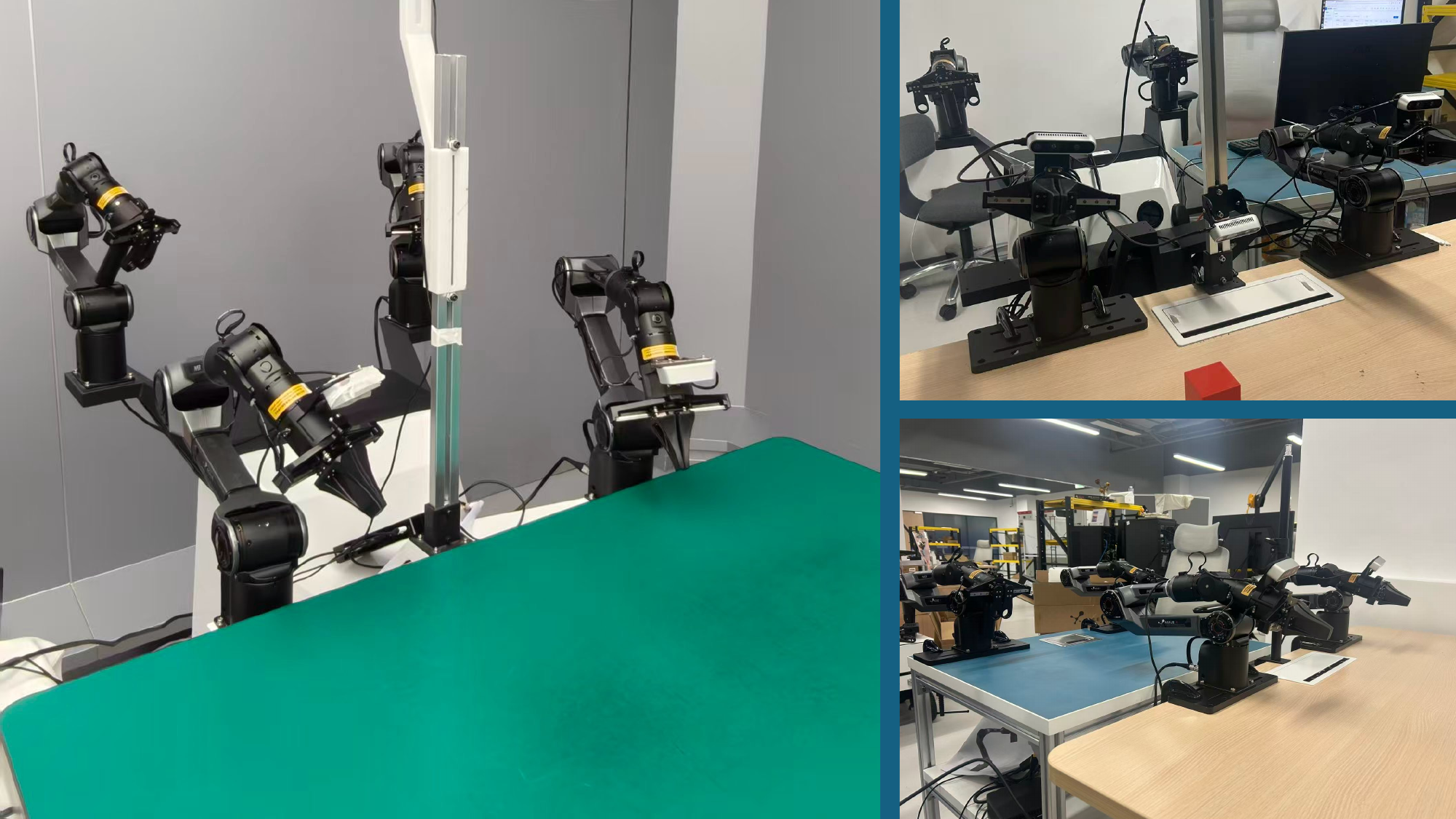}
    \caption{The demonstration showcases our backup hardware, which shares the same robot configurations and setup. This ensures that real‑world evaluations can be conducted reliably and at scale.}
    \label{fig:backup-hardware}
\end{figure}

\begin{table}[h!]
\centering
\renewcommand{\arraystretch}{1.3}
\caption{Summary of Task Descriptions, Steps, and Success Criteria}
\label{table:tasks}
\resizebox{1\textwidth}{!}{
\begin{tabular}{lll}
\hline
\textbf{Task Description} & \textbf{Steps} & \textbf{Success Criteria} \\
\hline
\multirow{3}{*}{\textbf{Click Bell}} &
(1) Move the arm toward the bell. & \multirow{3}{5cm}{The button has been successfully pressed within a limited number of action steps.} \\
& (2) Align and press the button. & \\
& (3) Return the arm to the initial position. & \\

\hline
\multirow{3}{*}{\textbf{Items Hand-Over and Place}} &
(1) Grasp the pen. & \multirow{3}{5cm}{The pen has been successfully placed in the brush pot within a limited number of action steps.} \\
& (2) Pass the pen to another arm. & \\
& (3) Place the pen in the brush pot. & \\

\hline
\multirow{3}{*}{\textbf{Dual-Arm Water Pouring}} &
(1) Grasp the water bottle and the cup. & \multirow{3}{5cm}{The water has been successfully poured into the cup, and both items are placed on the table within a limited number of action steps.} \\
& (2) Pour the water from the water bottle into the cup. & \\
& (3) Place both objects securely on the table. & \\

\hline
\multirow{3}{*}{\textbf{Table Rearrangement}} &
(1) Grasp the spoon and the fork. & \multirow{3}{5cm}{The spoons and forks have been arranged around the plate within a limited number of action steps.} \\
& (2) Arrange them neatly on both sides of the plate.\\ & \\

\hline
\multirow{4}{*}{\textbf{Basket Pick-and-Place}} &
(1) Grasp the milk box. & \multirow{4}{5cm}{The milk box has been placed into the basket, and the basket has been set at the table center within a limited number of action steps.} \\
& (2) Place the milk box in the basket. & \\
& (3) Grasp the basket. & \\
& (4) Place the basket in the center of the table. & \\

\hline
\multirow{5}{*}{\textbf{Drawer Open and Place}} &
(1) Grasp the handle. & \multirow{5}{5cm}{The item has been successfully placed in the drawer within a limited number of action steps.} \\
& (2) Open the drawer. & \\
& (3) Grasp an item. & \\
& (4) Place the item in the drawer. & \\
& (5) Close the drawer. & \\

\hline
\multirow{3}{*}{\textbf{Mixer Operating}} &
(1) Grasp the beaker. & \multirow{3}{5cm}{The beaker was placed on the mixer and stirring successfully started within a limited number of action steps.} \\
& (2) Place the beaker on the mixer. & \\
& (3) Start the mixer. & \\

\hline
\multirow{4}{*}{\textbf{Item Assembly}} &
(1) Grasp the silicone tubes. & \multirow{4}{5cm}{The two silicone tubes have been successfully joined together within a limited number of action steps.} \\
& (2) Adjust them to a horizontal position. & \\
& (3) Move one tube forward. & \\
& (4) Move the other tube and splice together. & \\

\hline
\multirow{4}{*}{\textbf{Manipulate Pipette}} &
(1) Grasp the pipette. & \multirow{4}{5cm}{The pipette has been correctly inserted into the beaker and the button has been successfully pressed within a limited number of action steps.} \\
& (2) Insert pipette into the first beaker. & \\
& (3) Another arm presses the pipette button.\\ & \\

\hline
\multirow{3}{*}{\textbf{Sample Loading}} &
(1) Grasp the test tube. & \multirow{3}{5cm}{The test tube has been successfully placed in the rack within a limited number of action steps.} \\
& (2) Pass the test tube to another arm. & \\
& (3) Place the test tube into the rack. & \\
\hline

\end{tabular}
}
\end{table}

% \newpage
\section{Evaluation Results of baselines}
\label{sec:eval-results}

This section presents the baseline evaluation results, which encompass performance comparisons across policies trained on pure simulation data, pure real-world data, as well as the $\pi_0$, $\pi_{0.5}$, and Motus model variants. The evaluations are conducted across three core dimensions: success rate, action steps, and inference time. 

Crucially, the empirical results show that the metrics of models trained on our simulation data are closely comparable to, and in several scenarios even outperform, those trained on real-world data when deployed in the real world. This demonstrates that our simulation synthesized data has comparable quality with real data, establishing a robust foundation for participants to conduct subsequent Sim2Real policy training and model optimization.
\end{document}